%% file: main.tex
\documentclass{IEEEtran}
\usepackage[utf8]{inputenc}

\usepackage{graphicx}
\usepackage{float}

\usepackage{amssymb}
\usepackage{amsmath}

\usepackage{booktabs, siunitx, tabularx}
\usepackage{subcaption}

\usepackage{authblk}

\usepackage{newtxmath}
\usepackage{bm}
\usepackage{adjustbox, lipsum}

\usepackage{dblfloatfix}

\title{Zero NeRF: Registration with Zero Overlap}

\let\oldtwocolumn\twocolumn
\renewcommand\twocolumn[1][]{%
    \oldtwocolumn[{#1}{
    \begin{center}
           \includegraphics[width=\textwidth]{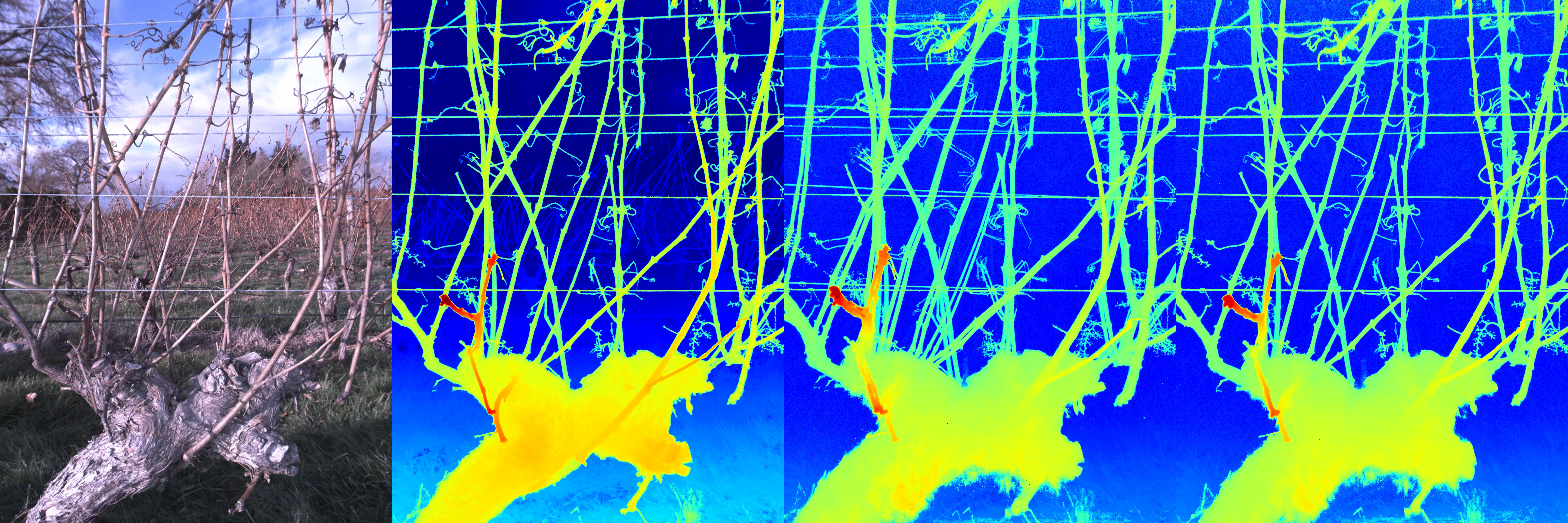}
           \captionof{figure}{Widely used registration methods such as RANSAC are ill-posed in the absence of visual correspondence between reconstructions. We present Zero-NeRF, the first method to our knowledge capable of solving this class of problems. (a) rendered image from NeRF, (b) rendered depth map from NeRF, (c) rendering of combined NeRF geometries using RANSAC alignment, (d) rendering of combined NeRF geometries using Zero-NeRF alignment.}
           \label{fig:title_banner}
        \end{center}
    }]
}

\author{Casey Peat, Oliver Batchelor, Richard Green, James Atlas}

\date{May 2022}

\begin{document}

\maketitle

\begin{abstract}

We present Zero-NeRF, a projective surface registration method that, to the best of our knowledge, offers the first general solution capable of alignment between scene representations with minimal or zero visual correspondence. To do this, we enforce consistency between visible surfaces of partial and complete reconstructions, which allows us to constrain occluded geometry. We use a NeRF as our surface representation and the NeRF rendering pipeline to perform this alignment. To demonstrate the efficacy of our method, we register real-world scenes from opposite sides with infinitesimal overlaps that cannot be accurately registered using prior methods, and we compare these results against widely used registration methods.

\end{abstract}

\input{tex/1_introduction}

\input{tex/2_related_work}

\input{tex/3_preliminaries}

\input{tex/4_method}

\input{tex/5_results}

\input{tex/6_conclusion}

\bibliographystyle{ieeetr}
\bibliography{references}


\end{document}

%% file: tex/1_introduction.tex
\section{Introduction}

In this work, we present Zero-NeRF, a method based on the recent advances in neural rendering that allow us to perform local registration between a pair of NeRF reconstructions with practically zero surface reconstruction overlap. To our knowledge, this is the first work to implement and demonstrate a method capable of solving this class of registration problems. In addition, we contribute a set of methods for performing NeRF reconstructions on real scans of thin objects and provide a dataset for registration of both real grapevine scans (shown above in Fig.~\ref{fig:title_banner}) and a synthetic one with known ground truth registration.

By framing the problem as a form of projective ICP between surfaces, bringing surfaces together while being constrained by the silhouette, surfaces can be almost perfectly registered with (almost) zero overlap when starting from an approximate solution. We measure surface depth for a given pixel/ray to quantify this error and minimise the difference between surface sets.

We demonstrate our method by registering Neural Radiance Field (NeRF) \cite{mildenhall2020nerf} reconstructions taken from opposite sides of a plant row. We visualise this in (Fig.\ref{fig:title_banner}c). Here, we can see a doubling effect due to visible occluded surfaces, which indicates an incorrect transform estimate. Using the NeRF rendering framework, the projective ICP is realized through the use of a differentiable depth estimation which we show converges fast and reliably on both real datasets and synthetic datasets with ground truth.

%% file: tex/2_related_work.tex
\section{Related Work}
\subsection{Scene Registration}

Registration is the process by which estimates the relative transform between two models. Most often studied (in the 3D case) as point cloud registration, where two or more partial scans of a scene are registered to form a larger, more complete scan. There are two main classes of registration algorithm; \emph{local} registration, which requires an approximate solution as initialisation, and \emph{global} registration, which does not and usually involves finding explicit correspondences. The registration used in this paper is a form of local registration.

Iterative Closest Point (ICP) \cite{924423} and variants \cite{Generalized-ICP, Yang_2016, low2004linear} are the predominant form of local registration. ICP is well known to fall into local minima and has problems with scenes which only partially overlap, even when these scenes register perfectly. ICP does not rely on explicit correspondences but does rely on the overlap; in contrast, we examine cases with (almost) zero surface overlap.

In terms of global registration, we make use of Random sample consensus (RANSAC) \cite{fischler1981random}, with Fast Point Feature Histograms (FPFH) \cite{rusu2009fast} features as a baseline, and to find a good starting point for local registration. Modern research in this area often focuses on matching features from deep learning \cite{yew20183dfeat, wang2019deep, ki2019pointnetlk} and also end-to-end learned methods \cite{lu2019deepvcp, horn2020deepclr}.

\subsubsection {Projective ICP}

Projective ICP was proposed in \cite{400574}, which minimises the distance between surfaces along a projected ray.  Various subsequent methods have used projections for registration, such as Kinect Fusion \cite{izadi2011kinectfusion}.

\subsubsection {Registration From Sihlouette}

Of the relevant literature, we found \cite{10.1007/978-3-319-16808-1_31} to be the closest to our method. This work uses a projective ICP scheme to register surface scans of objects from as few as four synthetic views of a turntable object with only a small amount of overlap. While their results are promising, they lack evaluation on either a published dataset or real scans. Here we demonstrate our solution on real scans and provide a quantitative evaluation using a synthetic dataset with ground truth registration.

\subsubsection {Agricultural Applications}

In the agricultural domain, \cite{Dong_2019} has explored this in the context of apple trees by exploiting the assumed cylindrical symmetry of tree (which isn't always true and is approximate) trunks and an attempt at using image silhouettes (which is only ever approximate as geometry is not perfectly planar and views not perfectly symmetrical. Both are domain-specific, whereas Zero NeRF is a general solution.

\subsection{Camera Pose Estimation}

Camera pose estimation typically involves either Structure from Motion (SfM) \cite{schonberger2016structure} or SLAM frameworks \cite{7219438, 4160954, 7747236}. Such methods operate by extracting and matching sets of sparse features \cite{790410} within a set of images and jointly optimising the 3D positions, camera poses and intrinsic camera parameters with a bundle adjustment. These methods crucially operate on sets of images of significant overlap to obtain feature correspondences between the images. 

It is common to employ SfM to register images and calibrate cameras as a prerequisite step to NeRF, as the camera poses (extrinsic) and intrinsic parameters are required as inputs. NeRF has also previously been used to directly fine-tune camera poses \cite {lin2021barf} and geometry together, using photometric losses. More recently, attempts even estimate intrinsic parameters too \cite{wang2021nerf}, or a complete calibration, estimating correspondences, too \cite{jeong2021self}. These methods are ill-posed to solve surface registration tasks with minimal or zero visual correspondence. This limitation can be attributed to reliance on the photometric loss used in the original NeRF method. Work such as NeRF-Pose \cite{li2022nerf} has explored the usefulness of NeRF model representations in pose estimation of highly occluded scenes. They constrain the problem space by formulating a dual loss that directly optimises for photometric consistently and for matching segmentation maps provided as a prior.

\subsection{Neural Scene Representations}

Recently NeRf \cite{mildenhall2020nerf} has shown that complex signals such as 5D light fields can be modelled implicitly via a trained multilayer perceptron. This idea builds on work such as \cite{NEURIPS2020_55053683} that use Fourier style encodings to overcome the inherent bias of neural networks towards low-frequency functions \cite{rahaman2019spectral}. Since then, work such as NeRf in the wild \cite{martin2021nerf} has added an image-specific latent embedding to model arbitrary non-consistent lighting conditions. Instant-NGP \cite{muller2022instant} has demonstrated significant efficiency gains by incorporating multi-resolution hashtables for feature storage and reducing the MLP depth and width, which now operate as decoders. Similarly, Plenoxels \cite{yu_and_fridovichkeil2021plenoxels} do away with neural networks entirely. They instead use sparse voxels to store the radiance field. MipNeRF-360 offers multiple contributions, such as a weight consolidation loss to bias NeRF optimisation towards producing unimodal visibility distributions and further constrain the inherently ill-posed problem of 2D-3D function estimates. Additionally, they use a coordinate system transform that maps unbounded Euclidean coordinates to a unit sphere and retains continuity and differentiability at all points within the coordinate system.

%% file: tex/3_preliminaries.tex
\section{Preliminaries}
Here we briefly describe the relevant aspects of the NeRF methodology and defer a full description to the original work \cite{mildenhall2020nerf}. Given a set of images $\{I_{i}\}^{N}_{i=1}$ with known camera parameters, NeRF learns to represent the scene as a 5D radiance field that we can be sample to obtain the opacity and radiance of a point observed from some direction. We can define NeRF with the following function $\mathcal{F}'_{\theta}(x, d) \rightarrow (\sigma, c)$ with $\theta$ representing the learned weights. For our alignment method, we are only interested in the opacity output $\sigma$, which is solely dependent on the sample position $x$, henceforth we simplify our NeRF function definition to $\mathcal{F}_{\theta}(x) \rightarrow \sigma$.

To render a pixel, NeRF projects it as a ray into the scene. We can represent this ray as a vector $r(t) = \mathbf{o} + t\mathbf{d}$ with $\mathbf{o}$ and $\textbf{d}$ being the origin and direction respectively, and t denotes distance along the ray. The ray is then sampled across a set of discrete distances $\{t_i\}^{N}_{i=1}$ to get a set of positions $\{x_i\}^{N}_{i=1}$ which act as a discrete approximation of our ray. A NeRF is then queried with the position and direction of these points, generating the density and radiance as outputs.

With the density distribution, we can translate this to opacity via numerical integration $\alpha_i = 1 - \exp(-\sigma_i \delta_i)$ and then apply (eq.~\ref{eq_weights}), which equates to recursive alpha compositing \cite{wallace1981merging}, to calculate the visibility weighting $w_{i}=\mathcal{W}(x_{i})$ relative to the pixel perspective. We can then calculate the pixel colour $\mathcal{C}$ as the dot product of the sample visibility weighting and radiance $\mathcal{C} = w \cdot c$, and calculate the pixel depth $\mathcal{D}_{Ne}$ as the dot product of the sample visibility weight and distance $\mathcal{D}_{Ne} = w \cdot t$. 

\begin{equation}
    \mathcal{W}(x_i) = \exp\left(-\sum^{i-1}_{j=1}\sigma_j \delta_j\right)(1 - \exp(-\sigma_i \delta_i))
    \label{eq_weights}
\end{equation}

%% file: tex/4_method.tex
\section{Method}
We first introduce our surface registration method, which is agnostic to the underlying surface representation. Second, we discuss how these ideas can be implemented using NeRF to extract scene surfaces. Finally, we describe our Zero NeRF implementation.

\subsection{Surface Reconstruction Properties}
We describe the inputs and assumptions required for our registration method. First, the reconstruction method reconstructs the set of surfaces within the line of view of a union of the sensor perspectives. We denote these surface sets as $ \{S_{tgt} \subset S\} $ and $ \{S_{src} \subset S\} $, these corresponding to the target and source reconstructions and subsets of the merged set of visible surfaces $ S $. The sensors view a finite depth range from $D_{min}$ to $D_{max}$ relative to the sensor origin.

We define the surface depth function $ \mathcal{D}(S, r) $ as the distance to the first intersection along a ray $ r $ with our surface set $ S $, or when there is no intersection, a constant $ \mathcal{D}_{max} $ (eq.~\ref{eq:surface_depth_fumction}), as we assume the visible surface lies beyond the surface space.

\begin{equation}
    \mathcal{D}'(S, r) = 
    \begin{cases} 
     \mathcal{D}(S, r)  & \mathcal{D}(S, r) \neq \varnothing \\
     \mathcal{D}_{max} & \mathcal{D}(S, r) = \varnothing \\
   \end{cases}
   \label{eq:surface_depth_fumction}
\end{equation}

\begin{figure}[ht]
    \centering
    \includegraphics[width=0.5\textwidth]{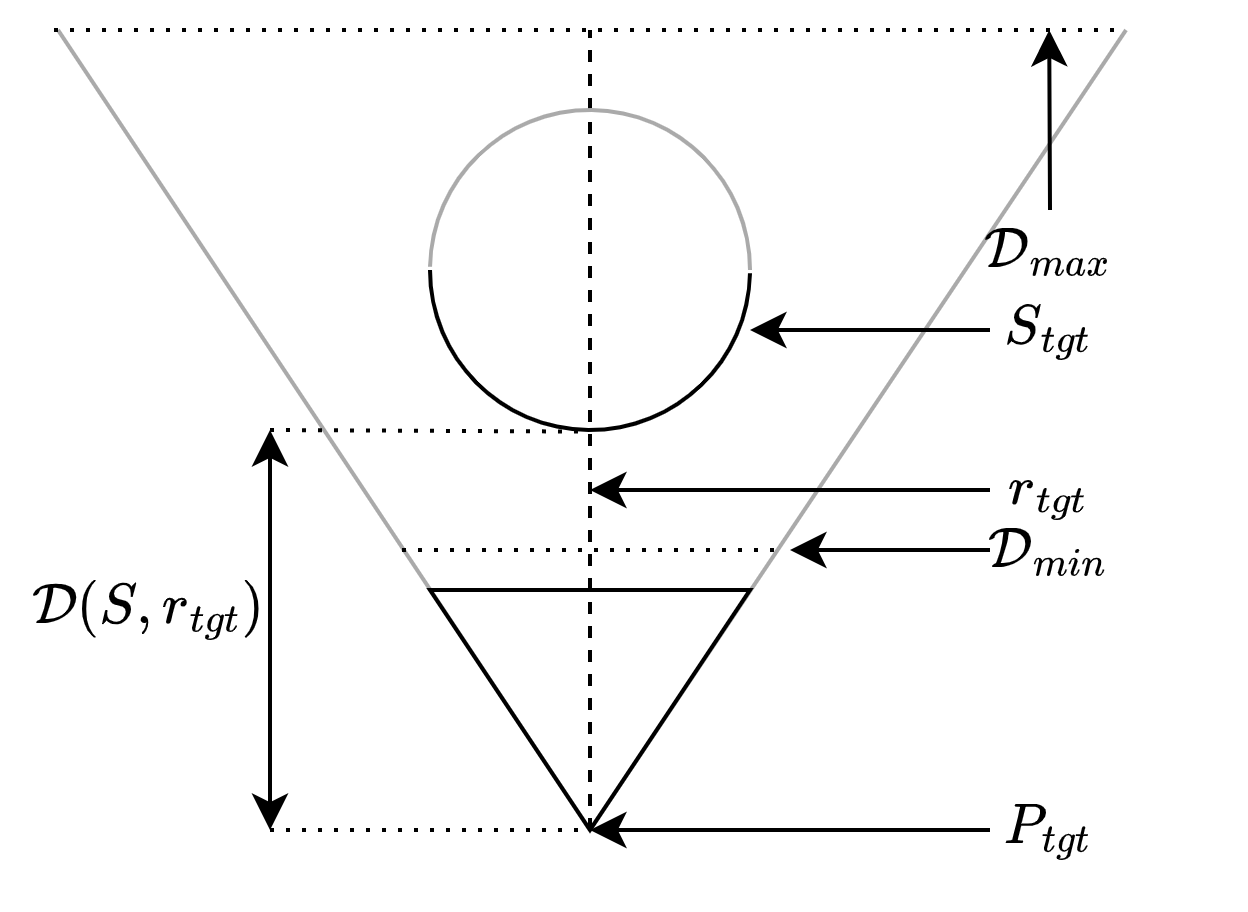}
    \caption{Surface depth function diagram of a ray intersecting the visible surface of a sphere}
    \label{fig:surface_depth_function}
\end{figure}

\subsection{Surface Registration}
\begin{figure*}[ht]
    \includegraphics[width=\textwidth]{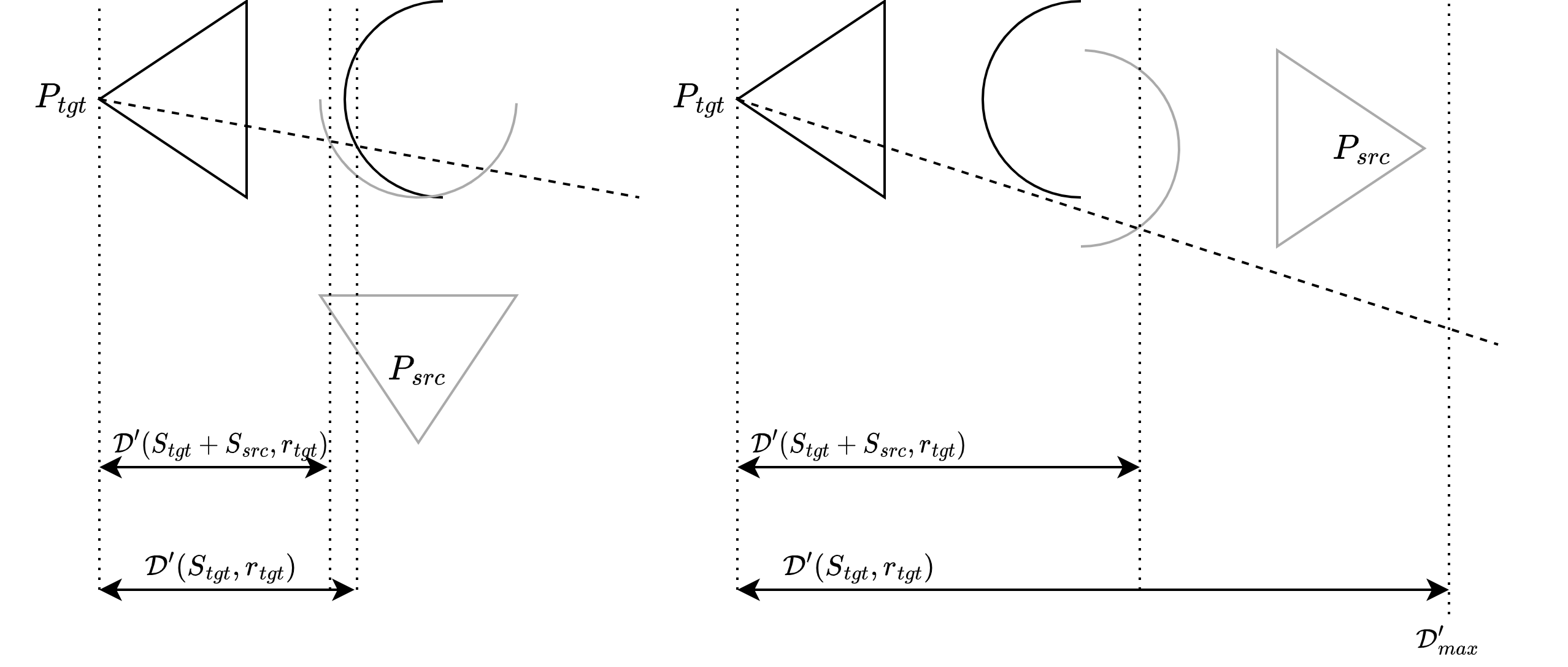}
    \caption{First example (2a) demonstrating depth error due to misalignment of a commonly visible point, Second example (2b) demonstrating error due to misalignment of a non-visible point relative to a target perspective}
    \label{fig:both_cases}
\end{figure*}

Our method aims to enforce consistency between the visible surfaces of our target surface reconstruction and the aggregate target and surface reconstruction when viewed from each target perspective. The relative transform between the target and source frame of references is unknown. Thus we denote a 6DoF transform $\mathbf{Z}$ as our estimate, which we wish to optimise. To measure this, calculate the difference between surface intersection depths for sample target ray $r_{tgt}$. We represent this with loss function (eq.~\ref{eq:depth_eq}), which we minimise via $\mathbf{Z}$.

\begin{equation}
    \mathcal{L}(\mathbf{Z}, r_{tgt}) = \mathcal{D}'(\mathcal{S}_{tgt}, r_{tgt}) - \mathcal{D}'\left(\mathcal{S}_{tgt} + \mathbf{Z}\mathcal{S}_{src}, r_{tgt}\right)
    \label{eq:depth_eq}
\end{equation}

To demonstrate this loss function, we visualise two misalignment cases. First, an example where the surface intersection occurs in both target $\mathcal{D}(S_{tgt}, r_{tgt}) \neq \varnothing$ and combined $\mathcal{D}(S_{tgt} + S_{src}, r_{tgt}) \neq \varnothing$ reconstructions (Fig.~\ref{fig:both_cases}a) due to misalignment of a commonly visible point. Second, where surface intersection does not occur in target $\mathcal{D}(S_{tgt}, r_{tgt}) = \varnothing$ however, it does in $\mathcal{D}(S_{tgt} + S_{src}, r_{tgt}) \neq \varnothing$ reconstructions (Fig.~\ref{fig:both_cases} due to the misalignment of a non-visible point relative to a target perspective).

\section{NeRF Implementation}

In this section, we describe our framework for implementing our method using NeRF as our scene representation and rendering framework. This process includes two stages, training our target and source NeRF as our target and source scene representations and estimating the relative transform between these NeRF reconstructions.

\subsection{NeRF}
We train a NeRF for each target and source image sets to function as our target and source scene representations. For the full description of this process, we defer to the original work [cite] and instead detail how our method differs. First, we use the coordinate system proposed by \cite{barron2021mip360} as this allows us to map an infinite euclidean volume into a finite cube using the following equation (eq.~\ref{eq:mipnerf_warping}). Additionally, we use the multi-resolution hash encoding method by \cite{muller2022instant}, which proposes storing learned features in a hash table and referencing by 3D location. This method speeds up training by orders of magnitude and makes real-time use a viable option. We use the learned latent vector for each image \cite{zhang2020nerf++} to model real-world lighting conditions. Finally, we use the weight consolidation loss proposed in \cite{barron2021mip360} which we propose the novel addition of annealing to improve its efficacy.

\begin{equation}
    x' = 
    \begin{cases} 
     x & |x| < 1 \\
     \left(2 - \frac{1}{|x|}\right)\left((\frac{x}{|x|}\right) & |x| > 1 \\
   \end{cases}
   \label{eq:mipnerf_warping}
\end{equation}

\subsection{Dynamic Weight Consolidation}

Given we are largely reconstructing surfaces rather than volumes, NeRF, in general, reconstructs a volumetric density function. The density along a ray takes on a continuous distribution which can lead to ambiguities as discussed in multiple works \cite{zhang2020nerf++}, as many different NeRFs can explain any arbitrary set of images without correctly representing the underlying geometry. 

In practice, the bias present in NeRF models prevents some of these problems, as certain scenes cannot be feasibly represented by a NeRF due to the information content of the high frequencies required in such a solution.  However, there are cases where the scene is not adequately constrained by the input perspectives, such as large homogeneous regions, an example being the sky and textureless surfaces.

\begin{figure}[ht]
    \includegraphics[width=\columnwidth]{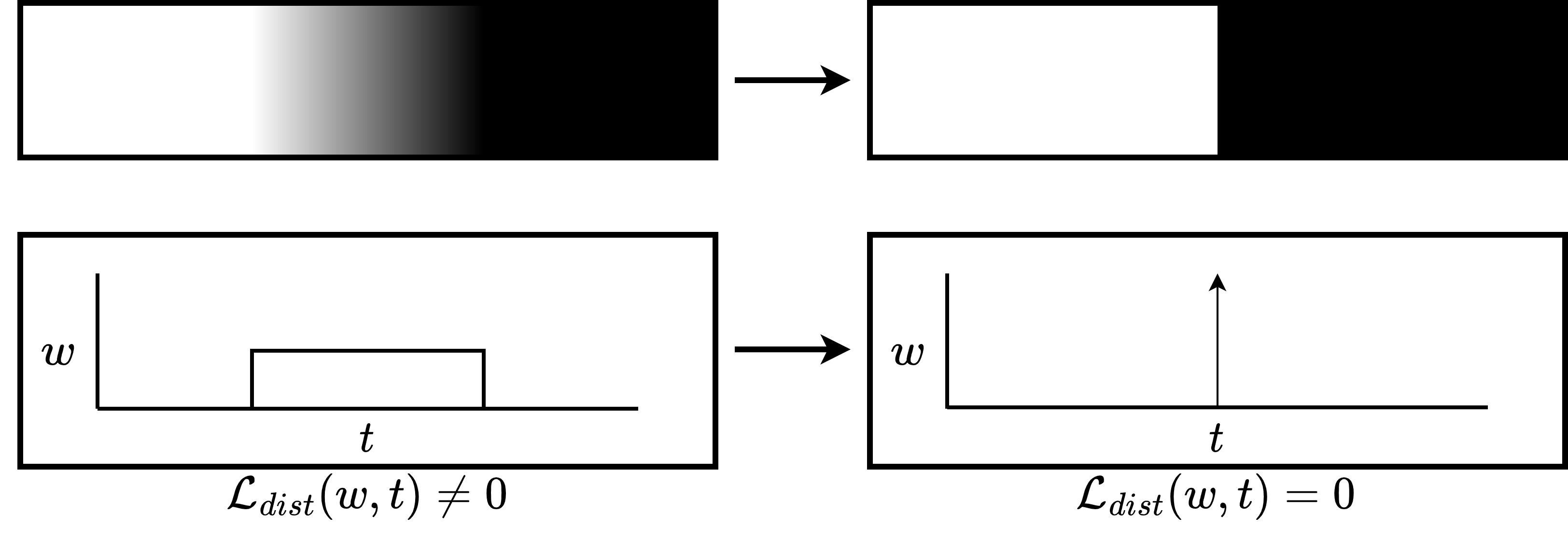}
    \caption{A visualisation of $\mathcal{L}_{dist}$ biasing NeRF representations towards defined surfaces}
    \label{fig:dist_loss}
\end{figure}

In \cite{barron2021mip360}, a prior termed weight consolidation is introduced on scene geometry (eq.~\ref{eq:weight_con}) to minimise the width of the sample weighting distribution, which we consider to be analogous to minimising the surface intersection of a ray from range, to a point (Fig.~\ref{fig:dist_loss}).

\begin{equation}
    \mathcal{L}_{dist}(w, t) = \sum^{N}_{i,j} w_i w_j |t_i - t_j|
    \label{eq:weight_con}
\end{equation}

\begin{equation}
    \mathcal{L} = \mathcal{L}_{photo} + \lambda \mathcal{L}_{dist}
\end{equation}

Our findings from experimenting with weight consolidation are that it does bias solutions towards more accurate surface representations. It was noticed when weighted too highly. This may prevent good convergence. We reason this is due to the continuous volume representation being advantageous early on, allowing a smooth transition to correct solutions. When the weight consolidation prior is applied right from the start of reconstruction, this interferes with convergence.

Instead of setting this weighting value as a constant hyperparameter, we anneal it using a dynamic weighting that increases logarithmically throughout training. This allows geometry to transition early on smoothly, then, as the training progresses, enforces that it converges on a surface representation.

\begin{equation}
    \lambda(i) = 10^{log_{10}(\lambda_2 \frac{i}{N}) + log_{10}(\lambda_1)}
\end{equation}

\section{Aligning NeRF models}
\begin{figure*}[ht]
    \centering
    \includegraphics[width=0.8\textwidth]{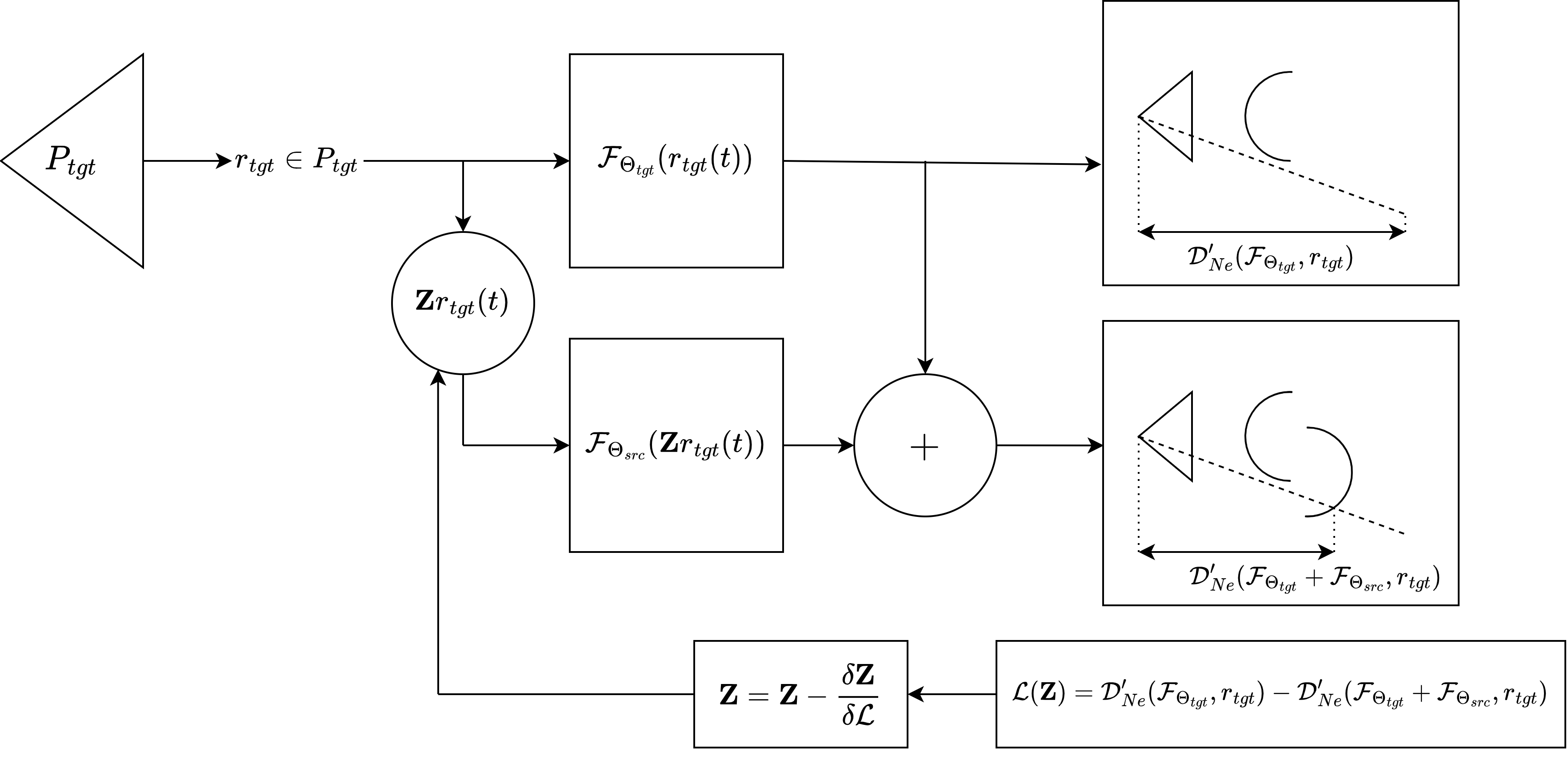}
    \caption{A flow diagram of our NeRF alignment optimisation procedure}
    \label{fig:transform_estimation_pipeline}
\end{figure*}

\begin{table*}[hb]
    \centering
    \begin{tabularx}{460pt}{ c || ccccc | ccccc }
        \toprule
        & \multicolumn{10}{c}{Pose Accuracy} \\
        Scene & \multicolumn{5}{c|}{Rotation Error (deg)$\downarrow$} & \multicolumn{5}{c}{Translation Error (mm)$\downarrow$} \\ \cmidrule{2-11}
              & RANSAC & RANSAC* & ICP & ICP* & Zero NeRF & RANSAC & RANSAC* & ICP & ICP* & Zero NeRF \\ \cmidrule{1-11}

        {Vine 0} & 0.4291 & 0.2543 & 0.0348 & 0.0446 & $\bm{0.0127}$ & 6.6821 & 11.3276 & 1.7580 & 2.9713 & $\bm{0.2189}$ \\
        {Vine 1} & 0.3737 & 0.3356 & $\bm{0.0143}$ & 0.0193 & 0.0631 & 6.2884 & 3.5301 & 1.8264 & 3.0980 & $\bm{0.4191}$ \\
        {Vine 2} & 0.6286 & 0.6961 & $\bm{0.0399}$ & 0.0837 & 0.0711 & 16.8421 & 13.5607 & 2.0065 & 4.1719 & $\bm{0.9811}$ \\
        {Vine 3} & 0.7741 & 0.2884 & 0.0816 & 0.1197 & $\bm{0.0649}$ & 4.9308 & 3.0945 & 2.7427 & 4.3661 & $\bm{1.1139}$ \\
        {Vine 4} & 2.6979 & 0.3741 & 0.0360 & 0.1193 & $\bm{0.0212}$ & 7.5545 & 6.2637 & 2.2825 & 3.0041 & $\bm{0.7054}$ \\
        {Vine 5} & 0.5250 & 0.5538 & $\bm{0.0246}$ & 0.0900 & 0.0401 & 10.2787 & 1.8157 & 1.0855 & 4.9095 & $\bm{0.7738}$ \\
        {Vine 6} & 0.1861 & 0.4348 & $\bm{0.0509}$ & 0.0711 & 0.0622 & 3.4213 & 7.9322 & 2.1515 & 4.6388 & $\bm{0.4260}$ \\
        {Vine 7} & -      & -      & 0.0833 & 0.1274 & $\bm{0.0127}$ & -      & -      & 2.5947 & 3.0852 & $\bm{0.2189}$ \\ \cmidrule{1-11}
        {Mean} & 0.8021 & 0.4195 & 0.0457 & 0.08436 & $\bm{0.0435}$ & 7.9997 & 6.7892 & 2.0560 & 3.7806 & $\bm{0.6071}$ \\ \bottomrule
    \end{tabularx}
\caption{A quantitative comparison of Zero-NeRF against alternative registration methods on synthetic data. Note that RANSAC failed to converge on "Vine 7". thus, ICP was initialised with perfect alignment, and Zero NeRF was initialised with the subsequent ICP alignment.}
\label{table:syntheticdata}
\end{table*}

With our NeRF models reconstructed, we now discuss our surface registration implementation. To add two NeRF geometries together, we add the density outputs of both target and source NeRF functions at the queried positions along a ray. As both functions operate under coordinate systems that differ by an unknown rigid transform, we transform the points of the ray using the transform $Z$, which approximates the target-to-source transform. We represent the output target densities as $\sigma$ (eq.\ref{eq:sigma_target}) and the output aggregate target and source density as $\hat\sigma$ (eq.\ref{eq:sigma_both})

\begin{equation}
    \sigma = \mathcal{F}_{\theta_{tgt}}(r_{tgt}(t))
    \label{eq:sigma_target}
\end{equation}

 \begin{equation}
    \hat\sigma = \mathcal{F}_{\theta_{tgt}}(r_{tgt}(t)) + \mathcal{F}_{\theta_{src}}(\mathbf{Z}r_{tgt}(t))
    \label{eq:sigma_both}
\end{equation}

As with the original NeRF work, we define transmittance $T_i(\sigma)$ (eq.~\ref{eq:transmittance}) as the probability that the ray has not terminated by sample $i$ given a set of densities along a ray $\sigma$.

\begin{equation}
    T_{i}(\sigma) = exp\left(-\sum^{i-1}_{j=1}\sigma_{j} \delta_{j}\right)
    \label{eq:transmittance}
\end{equation}

With this, we can define our depth-to-surface function with respect to NeRF geometric models $D_{Ne}$. We show the explicit derivation of this for a single NeRF representation (eq.~\ref{eq:dne_tgt}) and for the addition of NeRF representations (eq.~\ref{eq:dne_both}).

\begin{equation}
    \mathcal{D}_{Ne}(F_{\theta_{tgt}} , r_{tgt}) = \sum^{N}_{i=1}T_{i}(\sigma)(1-exp(-\sigma_{i} \delta_{i}))t_{i}
    \label{eq:dne_tgt}
\end{equation}

\begin{equation}
    \mathcal{D}_{Ne}(F_{\theta_{tgt}} + F_{\theta_{src}} , r_{tgt}) = \sum^{N}_{i=1}T_{i}(\hat\sigma)(1-exp(-\hat\sigma_{i} \delta_{i}))t_{i}
    \label{eq:dne_both}
\end{equation}

To ensure a valid depth in the case of a non-intersection, we weight the final depth with all remaining transmittance for the term $T_N(\sigma)t_N$ (eq.~\ref{eq:dmax_nerf}). This can be thought of as setting an infinite density at the final sample and thus guarantees ray termination within the defined space $ T_{N}(\sigma)(1-exp(-\infty \delta_{N}))t_{N} = T_{N}(\sigma)t_{N} $. This ensures a unit weighting sum across all depth samples, which is required to calculate the weighted average.

\begin{equation}
    \mathcal{D}'_{Ne}(F_{\theta_{tgt}} , r_{tgt}) = \sum^{N-1}_{i=1}T_{i}(\sigma)(1-exp(-\sigma_{i} \delta_{i}))t_{i} + T_{N}(\sigma)t_{N}
    \label{eq:dmax_nerf}
\end{equation}

With this, we have a NeRF implementation of our abstracted loss function (eq.~\ref{eq:l_dne}), which we minimise via gradient descent. We provide a diagram of this process shown in (Fig.~\ref{fig:transform_estimation_pipeline}).

\begin{equation}
    \mathcal{L}(\mathbf{Z}) = \mathcal{D}'_{Ne}(\mathcal{F}_{\theta_{tgt}} , r_{tgt}) - \mathcal{D}'_{Ne}(\mathcal{F}_{\theta_{tgt}} + \mathcal{F}_{\theta_{src}} , r_{tgt})
    \label{eq:l_dne}
\end{equation}

\begin{figure*}[ht]
    \centering
    \begin{subfigure}[b]{0.49\textwidth}
        \includegraphics[trim={0 8cm 0 8cm}, clip, width=\textwidth]{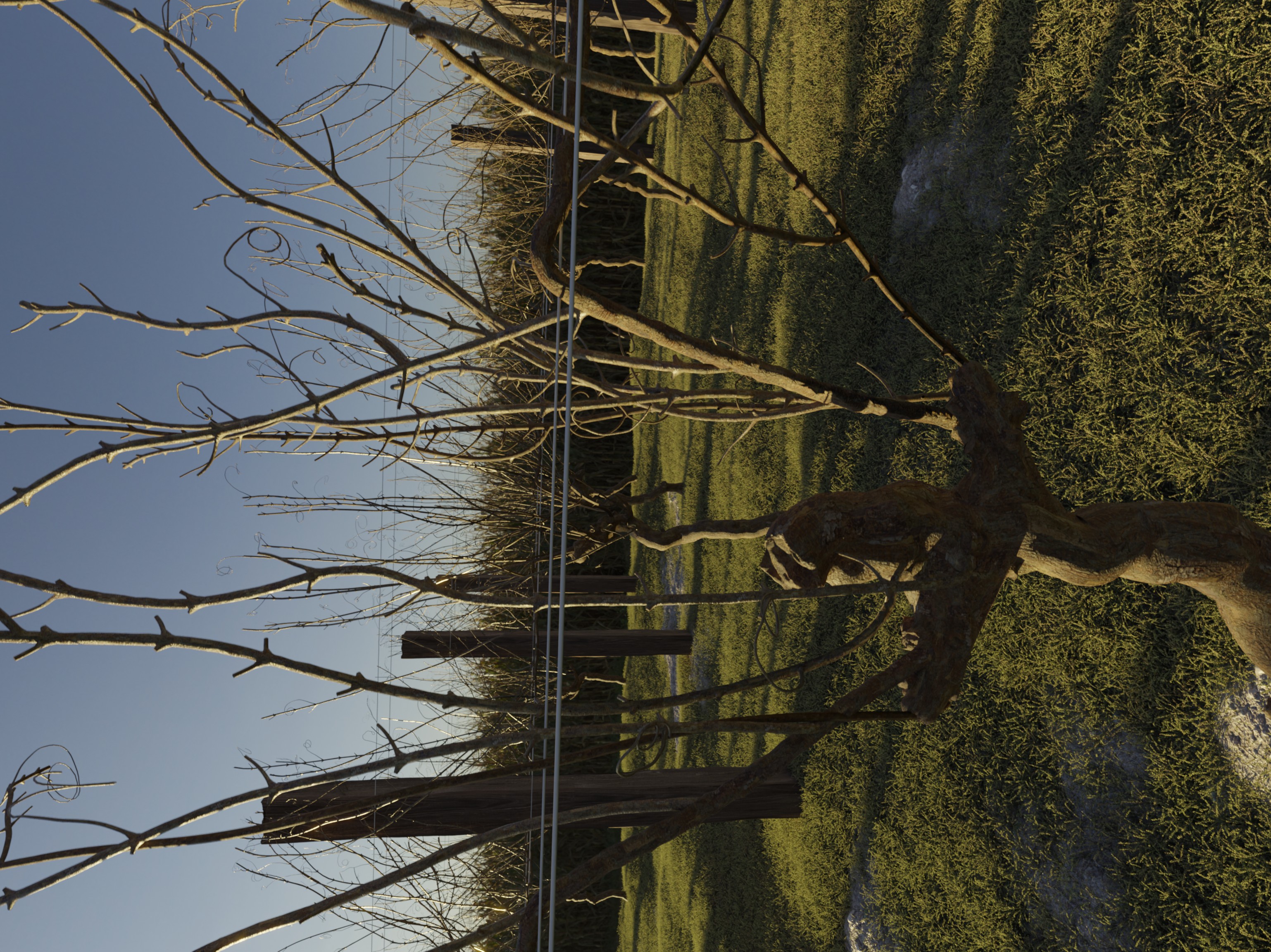}
        \caption{Camera perspective of a synthetic vine}
        \label{fig:dataset_example}
    \end{subfigure}
    \hfill
    \begin{subfigure}[b]{0.49\textwidth}
        \includegraphics[width=\textwidth]{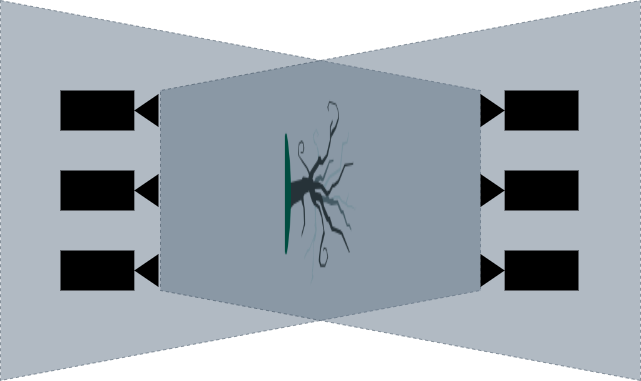}
        \caption{Front and rear camera poses relative to vine scene}
        \label{fig:scene_setup}
    \end{subfigure}
    \hfill
\end{figure*}

\begin{table*}[hb]
    \centering
    \begin{tabularx}{416pt}{ c || ccc | ccc | ccc }
        \toprule
        & \multicolumn{9}{c}{View Synthesis Quality} \\
        Scene & \multicolumn{3}{c|}{LPIPS$\downarrow$} & \multicolumn{3}{c|}{SSIM$\uparrow$} & \multicolumn{3}{c}{PSNR$\uparrow$} \\ \cmidrule{2-10}
              & RANSAC & ICP & Zero NeRF & RANSAC & ICP & Zero NeRF & RANSAC & ICP & Zero NeRF \\ \cmidrule{1-10}

        {Vine A} & 0.3646 & 0.3056 & $\bm{0.2907}$ & 0.7301 & 0.7765 & $\bm{0.7932}$ & 22.12 & 23.43 & $\bm{23.87}$ \\
        {Vine B} & 0.3608 & 0.3176 & $\bm{0.3002}$ & 0.7229 & 0.7620 & $\bm{0.7751}$ & 21.87 & 22.92 & $\bm{23.24}$ \\ 
        {Vine C} & 0.3475 & 0.3141 & $\bm{0.2983}$ & 0.7424 & 0.7696 & $\bm{0.7860}$ & 22.63 & 23.38 & $\bm{23.95}$ \\ \cmidrule{1-10}
        {Mean}   & 0.3576 & 0.3124 & $\bm{0.2964}$ & 0.7318 & 0.7694 & $\bm{0.7848}$ & 22.21 & 23.24 & $\bm{23.68}$ \\ \bottomrule
    \end{tabularx}
\caption{A quantitative comparison of Zero-NeRF against alternative registration methods on real-world scans.}
\label{table:realdata}
\end{table*}

%% file: tex/5_results.tex
\section{Results}
We evaluate our framework on our novel synthetic vines dataset (Fig.~\ref{fig:dataset_example}) consisting of 7 scenes, each of which consists of two sets of camera perspectives that image the front and back respectively (Fig.~\ref{fig:scene_setup}). Our 3D vine assets were created using speed tree \cite{SpeedTree} and rendered using blender \cite{Blender}. This dataset will be included along with the code release for this work. With this dataset, we highlight the problem of minimal common visible geometry and demonstrate the shortcomings of current surface registration research in these cases.

\subsection{Compared Methods}
We show our quantitative results in (Tab.~\ref{table:syntheticdata}), which compares our method Zero-NeRF with the Open3D \cite{Open3D} implementations of RANSAC (a global registration, based on FPFH features \cite{rusu2009fast}), and ICP using point clouds extracted from the nerf reconstructions. 

To demonstrate the significance of this registration problem, we also measure RANSAC and ICP using perfect point clouds generated from the ground truth depth maps. We denote these experiments with an *. From these perfect point cloud registration experiments, we can conclude that these registration methods are ill-posed, even assuming perfect reconstructions. Overall, our Zero NeRF surface registration method is sufficiently well posed to achieve sub-millimetre accuracy and significantly outperforms alternative surface registration methods in these cases. As ICP and Zero-NeRF are local refinement methods, we use RANSAC to perform the initial global pose estimation.

\subsection{Real Data}
To the best of our knowledge, our method is the first to accurately estimate the transformation between scene representations with minimal overlap. Thus we cannot compare against a ground-truth transform as a direct measurement of accuracy in a real environment. To measure performance, we train a new NeRF on the combined set of both source and target training images using our transform estimation and measure the rendering quality under said transform. Although NeRF can theoretically explain any arbitrary image set, function complexity increases significantly without a cohesive geometric representation. Hence we expect rendering accuracy to be correlated with transform estimation accuracy due to NeRF having limited parameters. From our results in (Tab.~\ref{table:realdata}), we can see that the transform estimation from Zero-Nerf results in the highest image quality metrics \cite{zhang2018unreasonable} \cite{wang2004image}. estimation from Zero-Nerf results in the highest image quality metrics \cite{zhang2018unreasonable} \cite{wang2004image}.

\begin{figure*}[ht]
    \centering
    \begin{subfigure}[b]{0.49\textwidth}
        \includegraphics[width=\textwidth]{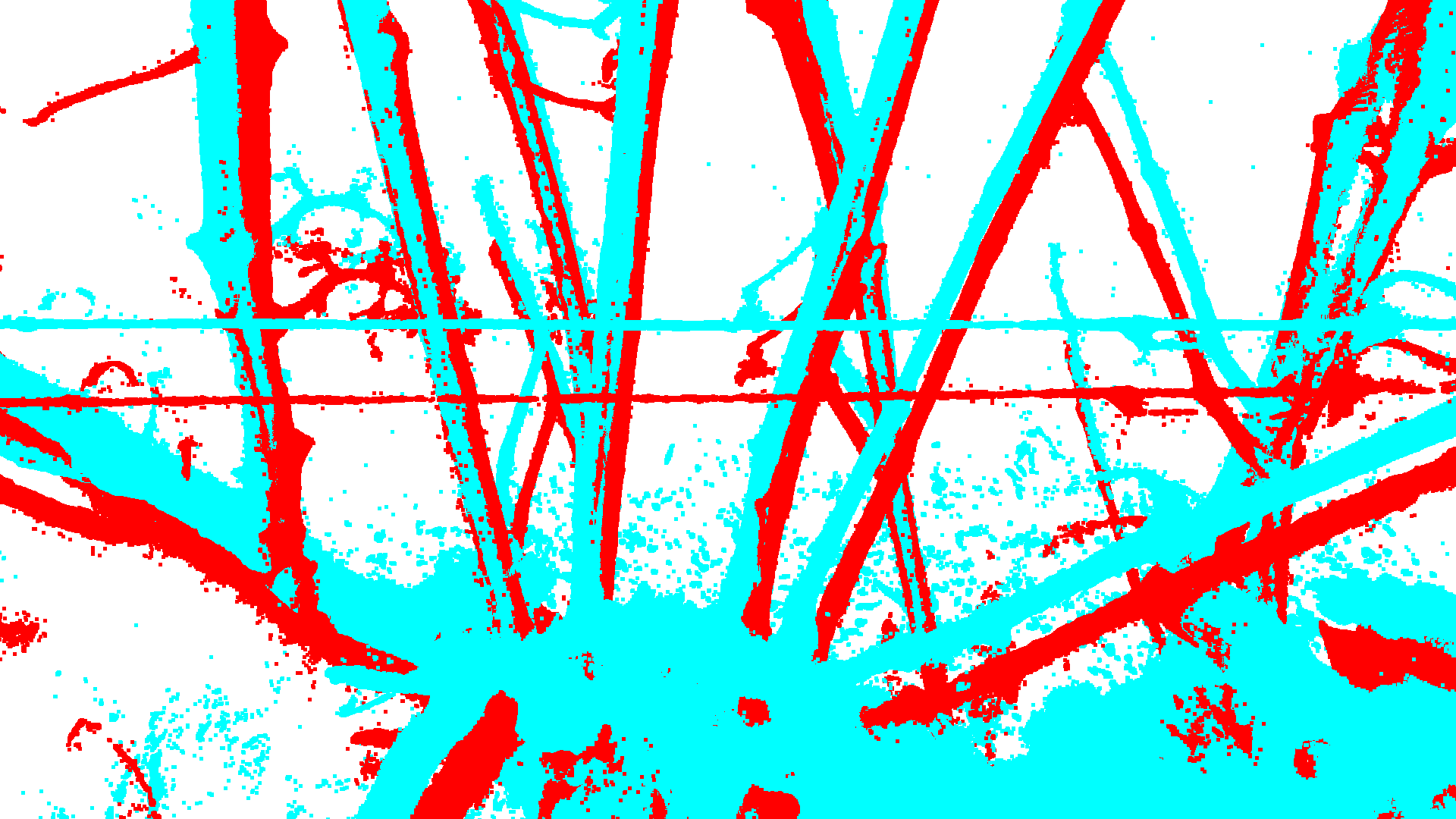}
        \caption{Pointclouds registered using RANSAC}
        \label{fig:ransac}
    \end{subfigure}
    \hfill
    \begin{subfigure}[b]{0.49\textwidth}
        \includegraphics[width=\textwidth]{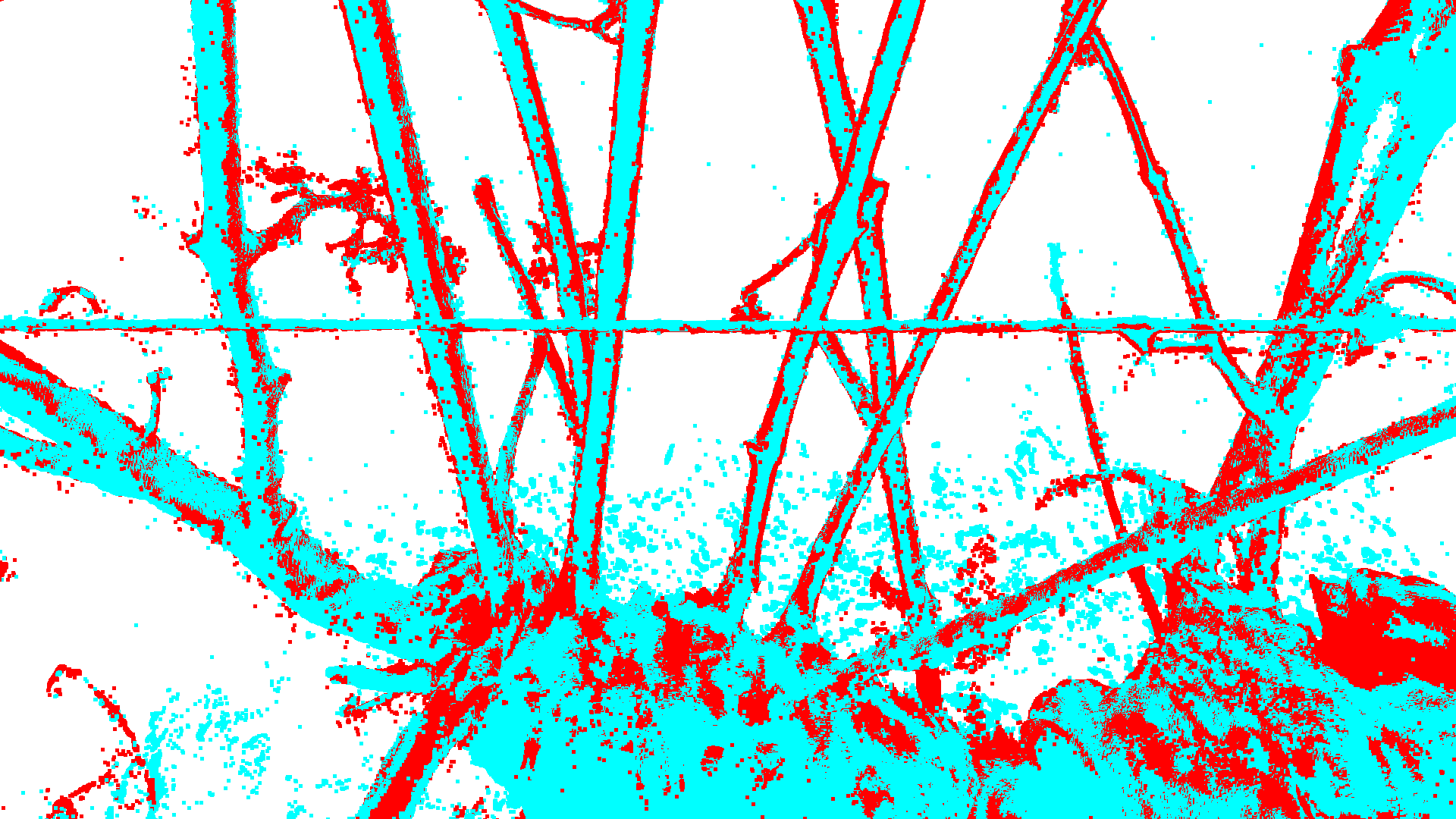}
        \caption{Pointclouds registered using Zero-Nerf}
        \label{fig:nerf}
    \end{subfigure}
    \hfill
\end{figure*}

We also show point cloud renderings of our results on real-world data. From this, we can see qualitatively that the wire is misplaced using RANSAC, as indicated by the doubling effect (Fig.~\ref{fig:ransac}). Using our Zero NeRF pose estimation, we see no such effect (Fig.~\ref{fig:nerf}).

\section{Experiment Details}
In these experiments, ZeroNeRF samples 4096 rays per iteration, 256 logarithmically spaced samples between 0.1 and 1 meters from the ray's origin, and 128 samples allocated using NeRF's hierarchical sampling procedure. We perform 20 epochs, each with 100 iterations. For gradient descent, we use the Adam optimiser. During training, we decrease the learning rate logarithmically from 5e-4 to 5e-6 for both the Euler rotation vector in radians and the translation vector in meters.

%% file: tex/6_conclusion.tex
\section{Limitations and future work}

One limitation of our work would be reliance on a global registration for initialisation. Fortunately, pose estimation via RANSAC worked in our real world datasets for this starting point. With poor overlap, however, reliable feature matches are not guaranteed. In practice, we have used this method to register scans of crops taken from both sides of a row. Instead of global registration, we instead achieve good local initialisation through localisation utilising Real Time Kinematic (RTK) GPS solutions.

The grapevine datasets we have used to evaluate this work involve many narrow structures. It is less clear if our method will be as efficient in cases with larger structures or where there is less visibility between the front and back sides of a scene; we intend to test this in future work of other agricultural settings such as apple tree orchards where occlusion between front and back is significantly greater due to leaves.

\section{Conclusion}
We propose Zero-NeRF, a NeRF-based framework that can accurately estimate the relative pose between two sets of images with zero visual correspondence. To achieve this, we offer the following contributions. First, we provide a novel mathematical framework to perform a surface registration agnostic to the specific reconstruction models. Second, we offer a means to implement this effectively using the NeRF rendering framework and demonstrate its efficacy compared with alternative registration methods. Third, we make an incremental improvement on the weight consolidation idea \cite{barron2021mip360} to promote discrete geometry while minimising interference to the training of NeRF. Our evaluations show that our method is state-of-the-art under this class of problems. To the best of our knowledge, it is the first method capable of solving these problems to a valuable degree for natural environments.

\section{Acknowledgements}
The research reported in this article was conducted as part of “MaaraTech: Data informed decision making and automation in orchards and vineyards”, which is funded by the New Zealand Ministry of Business, Innovation and Employment (UOAX1810).